\documentclass[letterpaper, 10 pt, conference]{ieeeconf} 
\IEEEoverridecommandlockouts                             

\usepackage{graphicx}
\usepackage{amsmath} 
\usepackage{amssymb}
\usepackage{booktabs}
\usepackage{xcolor}

\title{\LARGE \bf
DAISS: Phase-Aware Imitation Learning for Dual-Arm Robotic Ultrasound-Guided Interventions
}

\author{Feng Li$^{1,2*\dagger}$, Pei Liu$^{1*}$, Shiting Wang$^{1}$, Ning Wang$^{1}$, Zhongliang Jiang$^{3}$, Nassir Navab$^{1,2}$, and Yuan Bi$^{1,2}$ 
\thanks{*These authors contributed equally. $^{\dagger}$Corresponding author.}
\thanks{$^{1}$Chair for Computer Aided Medical Procedures(CAMP), Technical University of Munich(TUM), Munich, Germany}%
\thanks{$^{2}$Munich Center for Machine Learning (MCML), Munich, Germany}%
\thanks{$^{3}$The University of Hong Kong(HKU), Hong Kong, China}%
\thanks{Correspondence {\tt\small feng.li@tum.de}}%
}

\begin{document}

\maketitle
\thispagestyle{empty}
\pagestyle{empty}

\begin{abstract}
Imitation learning has shown strong potential for automating complex robotic manipulation. In medical robotics, ultrasound-guided needle insertion demands precise bimanual coordination, as clinicians must simultaneously manipulate an ultrasound probe to maintain an optimal acoustic view while steering an interventional needle. Automating this asymmetric workflow—and reliably transferring expert strategies to robots—remains highly challenging. In this paper, we present the Dual-Arm Interventional Surgical System (DAISS), a teleoperated platform that collects high-fidelity dual-arm demonstrations and learns a phase-aware imitation policy for ultrasound-guided interventions.  To avoid constraining the operator’s natural behavior, DAISS uses a flexible NDI-based leader interface for teleoperating two coordinated follower arms. To support robust execution under real-time ultrasound feedback, we develop a lightweight, data-efficient imitation policy. Specifically, the policy incorporates a phase-aware architecture and a dynamic mask loss tailored to asymmetric bimanual control. Conditioned on a planned trajectory, the network fuses real-time ultrasound with external visual observations to generate smooth, coordinated dual-arm motions. Experimental results show that DAISS can learn personalized expert strategies from limited demonstrations. Overall, these findings highlight the promise of phase-aware imitation-learning-driven dual-arm robots for improving precision and reducing cognitive workload in image-guided interventions.
\end{abstract}

\section{INTRODUCTION}

\begin{figure}[t]
    \vspace{0.5em}
    \centering
    \includegraphics[width=0.485\textwidth]{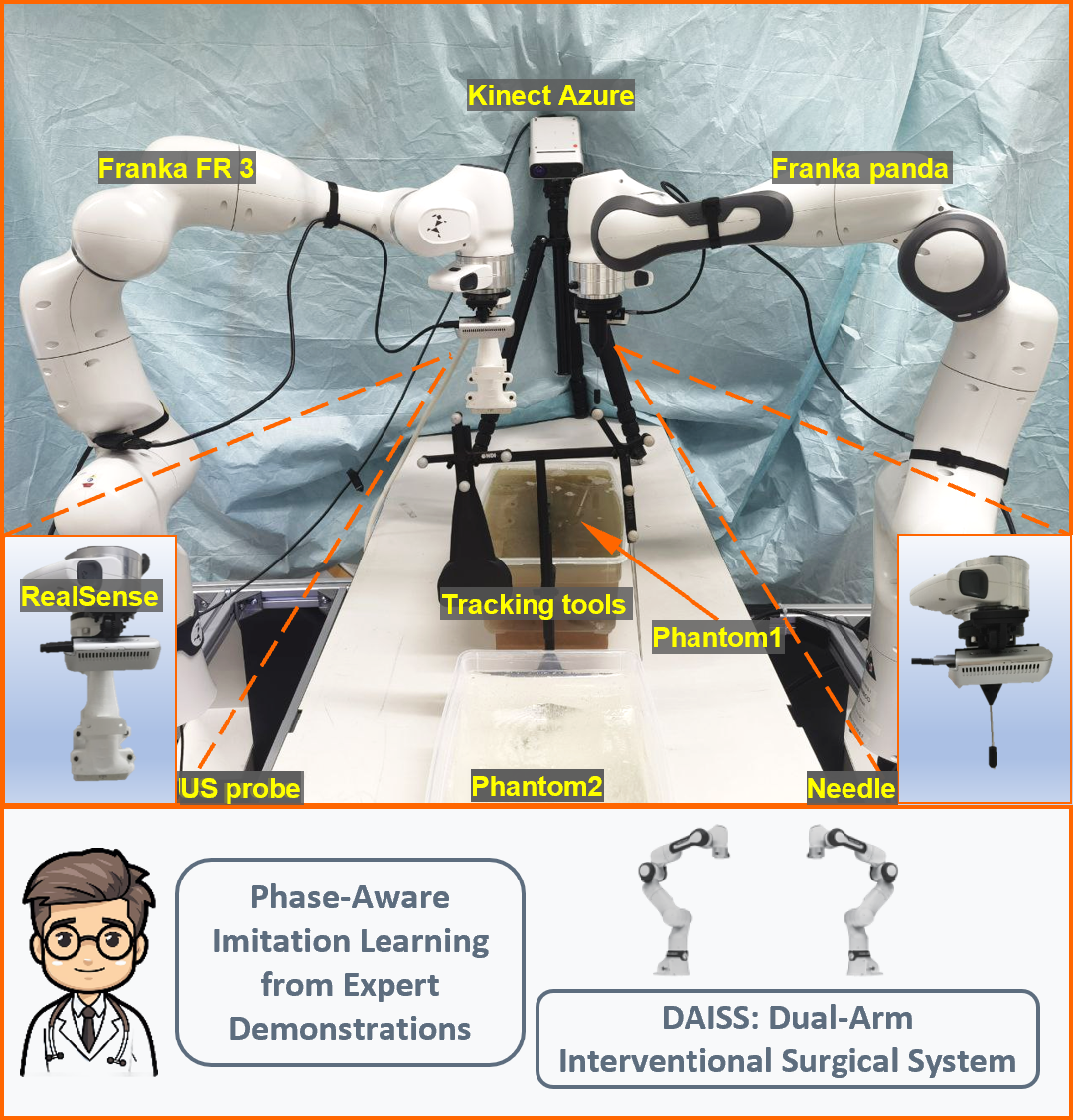}
     \caption{\textbf{Overview of the DAISS Platform and Phase-Aware Imitation Learning.} Our proposed bimanual teleoperation system integrates multimodal perception to execute complex ultrasound-guided interventions. By leveraging a dynamic phase-aware policy, DAISS effectively decouples the operation workflow, optimally balancing temporal efficiency with fine-grained kinematic precision.}
\label{fig:teaser}
\end{figure}

Ultrasound imaging is one of the most commonly used modalities in clinical practice due to its portability, absence of ionizing radiation, cost-effectiveness, and real-time imaging capability. Ultrasound-guided interventions are widely employed in needle-based procedures, including ablation, biopsy, vascular access, and regional anesthesia \cite{lorentzen2015efsumb}, and have become an integral part of minimally invasive treatments across multiple clinical specialties. In a typical ultrasound-guided procedure, the clinician manipulates an ultrasound probe with one hand while advancing an interventional instrument, such as a needle, with the other. During needle insertion, the operator must continuously interpret two-dimensional ultrasound images as visual feedback to mentally reconstruct three-dimensional anatomical information while simultaneously controlling the needle trajectory, which imposes substantial cognitive load and requires precise hand–eye coordination \cite{raitor2023design}. Achieving accurate needle placement, therefore, relies heavily on repeated training and clinical experience. In addition, procedural accuracy is often affected by physiological motion, such as respiration, tissue deformation, and involuntary hand tremor, particularly in high-precision interventions. Despite its widespread clinical adoption, ultrasound-guided treatment remains technically demanding and highly operator-dependent. These challenges motivate the development of robotic and intelligent systems aimed at stabilizing probe motion, improving needle–probe coordination, and reducing variability associated with manual operation.

Recent advances in robotic ultrasound have focused on supporting a broad range of clinical tasks by combining real-time ultrasound imaging with the precision and repeatability of robotic manipulation. By integrating imaging and positioning capabilities, robotic ultrasound systems enable applications such as three-dimensional (3D) volume compounding, autonomous scanning, and ultrasound-guided robotic intervention. These capabilities facilitate the fusion of ultrasound-derived volumes with other 3D imaging modalities, provide enhanced spatial context for clinical decision-making, and reduce operator workload \cite{li2021overview}. In the context of 3D volume acquisition, several studies have augmented robotic ultrasound systems with complementary sensing modalities to reconstruct accurate volumes, effectively mitigating motion artifacts and enhancing complex anatomical visualization \cite{jiang2021motion, li2025robotic}. For autonomous scanning, several studies have leveraged depth cameras to generate scanning paths, effectively providing visual perception for robotic ultrasound systems \cite{song2025intelligent, jiang2022towards}. Research on robotic ultrasound-guided intervention has progressed steadily over the past decade. Kojcev \emph{et al.} \cite{kojcev2016dual} introduced one of the first dual-arm systems for ultrasound-guided needle placement, integrating planning, imaging, and execution as a proof of concept for clinical translation. Subsequently, dual-arm robotic configurations—typically with one arm manipulating the ultrasound probe and the other controlling a needle—have been applied to a variety of clinical procedures, including central venous access \cite{koskinopoulou2023dual} and brachytherapy \cite{li2025ultrasound}. Despite these advances, most existing robotic ultrasound systems remain highly task-specific, relying on rigid control pipelines that exhibit limited adaptability across diverse patients and clinical scenarios. While 2D ultrasound provides real-time, targeted visualization of anatomical structures—allowing clinicians to quickly isolate specific lesions—its limited field of view inherently lacks broader spatial context. Consequently, navigating to the optimal imaging plane heavily relies on the operator's spatial awareness and intuitive decision-making, leading to a significant performance gap between experts and novices. Capturing these expert sensorimotor skills and translating them into robotic actions offers a promising solution to bridge this clinical gap. This motivation underscores the critical need for learning-based approaches that empower robotic systems to perceive, adapt, and act directly upon ultrasound data, paving the way for more intelligent, consistent, and generalizable interventions.

In the robotics community, a paradigm shift toward robot learning and data-driven policies has recently bridged the gap between task-specific automation and broad generalization. Zhao \emph{et al.} pioneered this with Action Chunking with Transformers (ACT) and the ALOHA platform \cite{zhao2023learning, zhao2024aloha}, demonstrating that fine-grained bimanual manipulation can be learned end-to-end on cost-effective hardware. Inspired by this success, the medical domain has begun to adopt this transformer-based paradigm for complex interventions. Kim \emph{et al.} proposed the Surgical Robot Transformer (SRT) \cite{kim2024surgical}, validating that such policies can master intricate soft-tissue manipulation tasks, while the subsequent SRT-H \cite{kim2025srt} introduced a hierarchical language-conditioned framework to autonomously handle long-horizon surgical procedures. In the context of the ultrasound, Jiang \emph{et al.} \cite{jiang2024intelligent} developed an intelligent robotic sonographer capable of autonomously exploring target anatomies and navigating the ultrasound probe to acquire standard imaging planes through expert demonstration learning. UltraDP \cite{chen2025ultradp} has recently applied diffusion policies to autonomous scanning. However, the application of bimanual imitation learning for coordinating a probe-holding arm and a needle-insertion arm remains underexplored. Furthermore, this coordination is inherently asymmetric and sequential: needle advancement typically occurs only after the ultrasound probe has secured and stabilized an optimal acoustic window. Consequently, an effective learning policy must be capable of dynamically shifting its attention and decoupling the control objectives for each arm to accommodate these distinct procedural phases.

To address these challenges, we present the Dual-Arm Interventional Surgical System (DAISS), a teleoperated platform that leverages phase-aware imitation learning for ultrasound-guided interventions. DAISS transfers expert bimanual strategies—maintaining optimal ultrasound views while advancing a needle—to a dual-arm robot. A flexible NDI-based leader interface enables intuitive demonstration collection, while a data-efficient learning policy executes robust insertions under real-time ultrasound feedback. The main contributions are threefold:

\begin{enumerate}
    \item We develop a dual-arm teleoperation platform with NDI-based leader--follower mapping and synchronized multimodal sensing (ultrasound, external vision, and robot states), providing data-efficient, high-fidelity demonstrations for learning clinically relevant probe--needle coordination behaviors.
    \item We propose a phase-aware imitation framework for ultrasound-guided intervention that explicitly models the sequential and asymmetric roles of the probe and needle arms. The policy couples phase conditioning with a dynamic mask loss and dual decoupled action heads, enabling stable coarse-to-fine control and improved coordination across procedural stages.
    \item We establish a mirrored dual-phantom evaluation paradigm to preserve kinesthetic consistency between demonstration and execution, and conduct extensive experiments showing robust policy transfer from limited demonstrations, accurate targeting, and stable ultrasound-view maintenance during intervention.
\end{enumerate}

\section{METHODS}

\begin{figure*}[!t]
    \centering
    \includegraphics[width=0.95\textwidth]{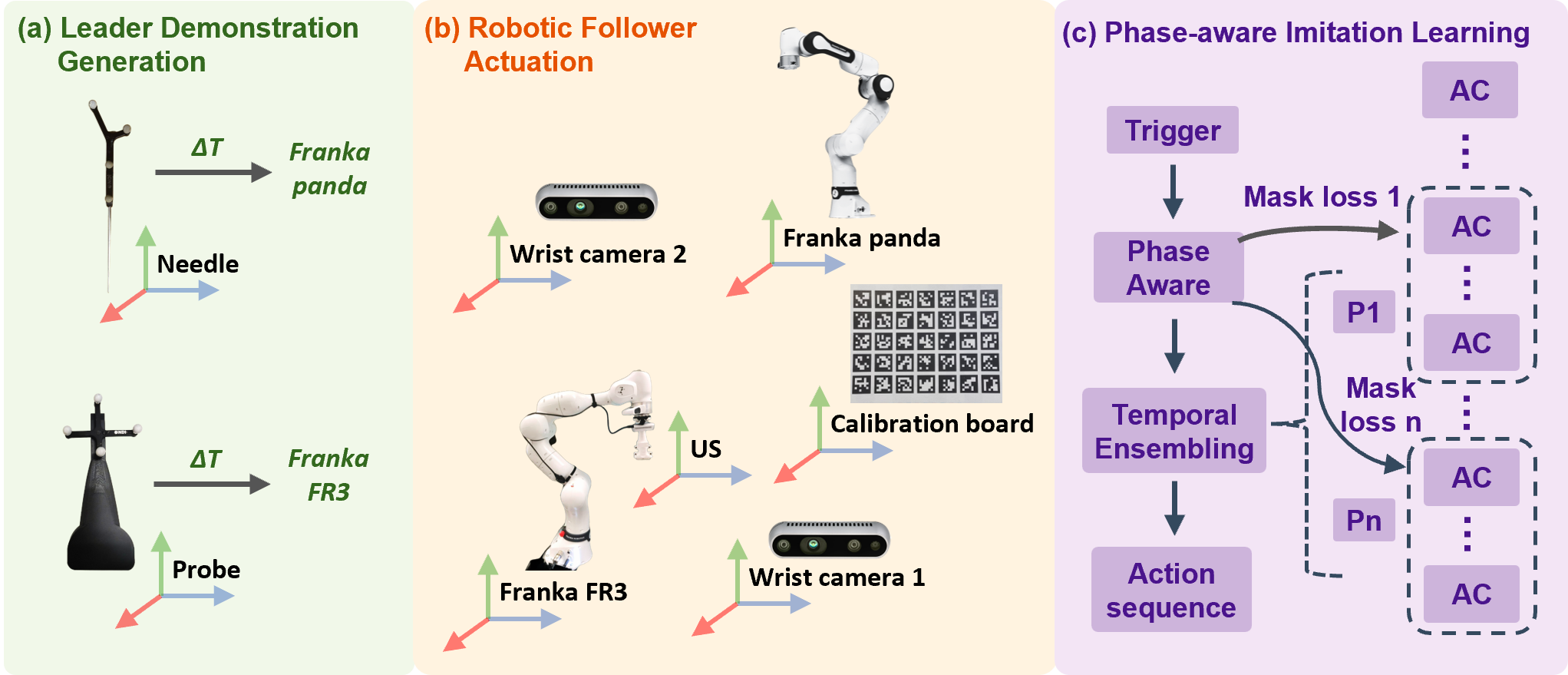}
      \caption{\textbf{The DAISS Framework.} (a) \textbf{Leader Demonstration Generation:} captures expert multimodal trajectories. (b) \textbf{Robotic Follower Actuation:} ensures safe, physical execution of the bimanual clinical workflow. (c) \textbf{Phase-Aware Imitation Learning:} maps the demonstrations to autonomous policies, employing a dynamic mask loss to resolve the speed-accuracy trade-off across different interventional phases. }
    \label{fig:overview}
\end{figure*}

\subsection{System Overview}

As illustrated in Fig.~\ref{fig:teaser}, the hardware architecture of DAISS comprises a leader--follower teleoperation setup. The robotic execution system features dual manipulators equipped with two wrist-mounted cameras and a global camera for full workspace observation. On the leader side, a custom-designed stand supports two optically tracked bimanual instruments: a simulated ultrasound probe and a simulated interventional needle. By leveraging an NDI tracking system to capture the real-time poses of these tracked tools, the framework ensures smooth, low-latency kinematic mapping to the robotic arms. Crucially, because optical pose tracking is inherently non-contact and thus cannot provide active force feedback, operators may experience reduced kinesthetic cues during telemanipulation. To mitigate this limitation, we introduce a physical dual-phantom paradigm. Two identical tissue phantoms, cast from the same mold with matching anatomical surface profiles, are deployed in the operator's and the robot's workspaces, respectively. This mirrored setup enables realistic haptic feedback during manipulation, thereby improving kinesthetic consistency between human demonstrations and robotic executions without relying on active force-feedback hardware.

Building upon this high-fidelity data collection framework, our learning policy is designed to handle long-horizon dual-arm interventions with phase-dependent precision requirements. Long-horizon operation procedures typically consist of distinct operational phases with varying precision demands. For instance, initial gross motions, such as navigating the ultrasound probe to the tissue surface, are relatively repeatable and exhibit lower variability, and can be learned as coarse trajectories. Conversely, critical fine manipulation tasks---such as dynamically adjusting the probe to secure an optimal acoustic window or precisely steering the needle to the target---are highly expert-dependent and require the network to capture fine-grained sensorimotor details. To effectively address these asymmetric, phase-wise learning objectives, we introduce a phase-aware network architecture, as shown in Fig.~\ref{fig:overview}(c). By explicitly differentiating between procedural phases, this structure conditions the policy on the current phase and encourages phase-specific feature specialization, promoting robust generalization during routine movements while better preserving nuanced expert behaviors during critical interventional steps.

\subsection{Hardware Architecture and Setup}
The hardware configuration of the DAISS platform integrates clinical imaging, bimanual actuation, and multimodal perception. The follower execution system comprises two collaborative manipulators: a Franka Research 3 (FR3, Franka Emika GmbH, Munich, Germany) for manipulating a clinical ultrasound probe (ACUSON Juniper, Siemens Healthineers AG, Erlangen, Germany), and a Franka Panda (Franka Emika GmbH, Munich, Germany) for interventional needle insertion. To capture rich visual context, two wrist-mounted RGB-D cameras (Intel RealSense D435, Intel Corporation, Santa Clara, CA, USA) are mounted on the arms to observe end-effector--tissue interactions in close range. This local vision is complemented by an overhead RGB-D camera (Azure Kinect DK, Microsoft Corporation, Redmond, WA, USA) that provides a global view of the intervention workspace. On the leader side, an optical tracking system (Polaris Vicra, Northern Digital Inc., Waterloo, ON, Canada) captures the real-time poses of two physician-operated leader instruments, each a high-accuracy geometric replica of the ultrasound probe or needle and equipped with optical markers.

To ensure accurate motion tracking, a 3D-printed stand serves as the docking station for the two leader tools, establishing a well-defined zero-reference pose. Simultaneously, the follower robotic arms are set to a matching initial configuration. By measuring the incremental 6-DoF pose changes relative to this reference, the NDI system continuously maps the operator's motions to the robots, enabling low-latency bimanual teleoperation. Crucially, this setup reduces the need for explicit cross-frame calibration between the leader and follower workspaces. By designing the tracking tools as high-accuracy geometric replicas of the ultrasound probe and needle, we align the distance from the tracking marker to the tool tip with the distance from the robot end-effector to the actual instrument tip. Thus, a direct incremental kinematic mapping can be applied with minimal additional calibration. Finally, to complete the setup, the structurally identical twin phantoms are manually aligned to match the working-surface height and minimize residual vertical offsets between the leader and follower environments.

\subsection{Kinematic Control}
Formally, to avoid unintended translation--rotation coupling and drift caused by incremental integration during rigid-body transformations, the kinematic mapping is decoupled into positional and rotational updates based on an absolute (reference-based) displacement mapping strategy. Let the robot's initial zero-reference end-effector state at the engagement phase be defined by its position $\mathbf{p}^{(0)}$ and rotation matrix $\mathbf{R}^{(0)}$ in the base frame. During teleoperation, the optical tracking system continuously captures the leader tool's movements. We compute the absolute spatial displacement $\Delta \mathbf{p}$ and rotational deviation $\Delta \mathbf{R}$ with respect to the leader tool's engagement pose. Instead of accumulating frame-to-frame errors, these transformations are mapped directly onto the robot's predefined zero-reference pose. The target end-effector pose at any given time step $t$ is then computed as:

\begin{equation}
\mathbf{p}^{(t)} = \mathbf{p}^{(0)} + \Delta \mathbf{p}, \quad \mathbf{R}^{(t)} = \Delta \mathbf{R} \mathbf{R}^{(0)}
\end{equation}

as shown in Fig.~\ref{fig:overview}(a). By continuously applying these decoupled reference-based transformations, the system robustly translates human intentions into precise robotic workspace coordinates while mitigating drift accumulation.

Raw positional data acquired from optical trackers often contain high-frequency jitter, which can cause erratic robotic joint velocities. To address this, we implement a dynamic target queue that buffers incoming absolute commands and generates smooth trajectory transitions at the robot control rate. For two consecutive target states, the translational component is interpolated using linear interpolation (LERP), while the rotational component is smoothed using Spherical Linear Interpolation (SLERP) to preserve constant angular velocity. To ensure the shortest rotation path and prevent erratic rotations, we strictly enforce $\mathbf{q}_{\text{prev}} \cdot \mathbf{q}_{\text{target}} \ge 0$ (negating $\mathbf{q}_{\text{target}}$ if necessary) to maintain quaternion hemisphere consistency before applying:

\begin{equation}
\mathbf{q}(\alpha) = \frac{\sin((1-\alpha)\theta)}{\sin(\theta)} \mathbf{q}_{\text{prev}} + \frac{\sin(\alpha\theta)}{\sin(\theta)} \mathbf{q}_{\text{target}}
\end{equation}

where $\alpha \in [0, 1]$ is the interpolation step, $\mathbf{q}$ denotes the unit quaternion representation of the rotation, and $\theta = \cos^{-1}(\mathbf{q}_{\text{prev}} \cdot \mathbf{q}_{\text{target}})$. This dual-interpolation scheme serves as a trajectory-smoothing mechanism, ensuring that the interventional tools move fluidly without sudden jerks.

\subsection{Bimanual Safety Mechanisms}

To ensure safe bimanual operation during cooperative manipulation, we implement a two-level safety mechanism. The first level enforces per-arm Cartesian workspace constraints in the robot base frame. Translational limits are defined along the three spatial axes; if a commanded teleoperation motion exceeds these bounds, the trajectory is saturated (clamped) to confine the end-effector within its allowable workspace. To facilitate the second level of safety---active inter-robot collision avoidance---a unified global coordinate system must first be established. We perform a standard hand-eye calibration \cite{horaud1995hand} for each manipulator using a shared calibration board, as depicted in Fig.~\ref{fig:overview}(b). This procedure determines the fixed transformation matrices between each wrist camera and its corresponding end-effector, denoted as ${^{e}}\mathbf{T}_{c}^{\text{FR3}}$ and ${^{e}}\mathbf{T}_{c}^{\text{Pan}}$. Consequently, the relative base transformation from the Panda to the FR3 can be formulated as:
\begin{equation}
\begin{split}
^{\text{FR3}}\mathbf{T}_{\text{Pan}} = {^{b}}\mathbf{T}_{e}^{\text{FR3}} \cdot {^{e}}\mathbf{T}_{c}^{\text{FR3}} \cdot {^{c}}\mathbf{T}_{m}^{\text{FR3}} \\ \left( {^{c}}\mathbf{T}_{m}^{\text{Pan}} \right)^{-1} \cdot \left( {^{e}}\mathbf{T}_{c}^{\text{Pan}} \right)^{-1} \cdot \left( {^{b}}\mathbf{T}_{e}^{\text{Pan}} \right)^{-1},
\end{split}
\end{equation}

where ${^{b}}\mathbf{T}_{e}$ denotes the forward kinematics transformation from the robot base to its end-effector (directly accessible via the robot controller), and ${^{c}}\mathbf{T}_{m}$ represents the relative pose of the calibration board with respect to the camera (estimated through visual tracking) for each arm. This calibration allows both robots and tools to be expressed in a common frame for real-time collision checking.

Building upon this shared spatial awareness, the second safety level actively prevents physical interference within the constrained shared workspace. Performing high-frequency collision detection on complex 3D robotic meshes is computationally expensive and can introduce unacceptable latency into the teleoperation pipeline. To address this, we geometrically abstract the distal segments of the robots---including the ultrasound probe and the interventional needle---as virtual spatial cylinders. Each bounding cylinder is defined by a central line segment and a physical bounding radius $r$, where the centerline is used for distance computation and $r$ accounts for tool thickness. Let $\mathbf{S}_1(\lambda_1) = \mathbf{a}_0 + \lambda_1 \mathbf{u}$ and $\mathbf{S}_2(\lambda_2) = \mathbf{b}_0 + \lambda_2 \mathbf{v}$ denote the central axes representing the two instruments in the unified global frame. Here, $\mathbf{a}_0$ and $\mathbf{b}_0 \in \mathbb{R}^3$ are the 3D position vectors indicating the end-effector attachment points, while $\mathbf{u}$ and $\mathbf{v}$ represent the normalized directional vectors pointing along the shafts of the respective instruments. The scalar multipliers $\lambda_1, \lambda_2 \in [0, L_1] \ \text{and} \ [0, L_2]$ parameterize the physical lengths of the tools. This geometric abstraction reduces the complex collision detection problem into a highly efficient segment-to-segment distance calculation. The system continuously computes the minimum Euclidean distance $d_{\text{min}}$ between the two centerlines:

\begin{equation}
d_{\text{min}} = \min_{\lambda_1, \lambda_2} \| \mathbf{S}_1(\lambda_1) - \mathbf{S}_2(\lambda_2) \|_2
\end{equation}

Because $d_{\text{min}}$ only measures the distance between the centerlines, a commanded trajectory is deemed safe and executed only if $d_{\text{min}} \geq 2r + c$. The term $2r$ accounts for the physical volume of the tools (assuming symmetric radii for the bounding cylinders), and $c$ acts as a designated safety clearance margin. Any kinematic command violating this threshold is blocked (and the system maintains the last safe target), thereby preventing inter-robot collisions with minimal computational overhead.

\subsection{Phase-aware Imitation Learning Framework}
\begin{figure}
\centering
\includegraphics[width=0.45\textwidth]
{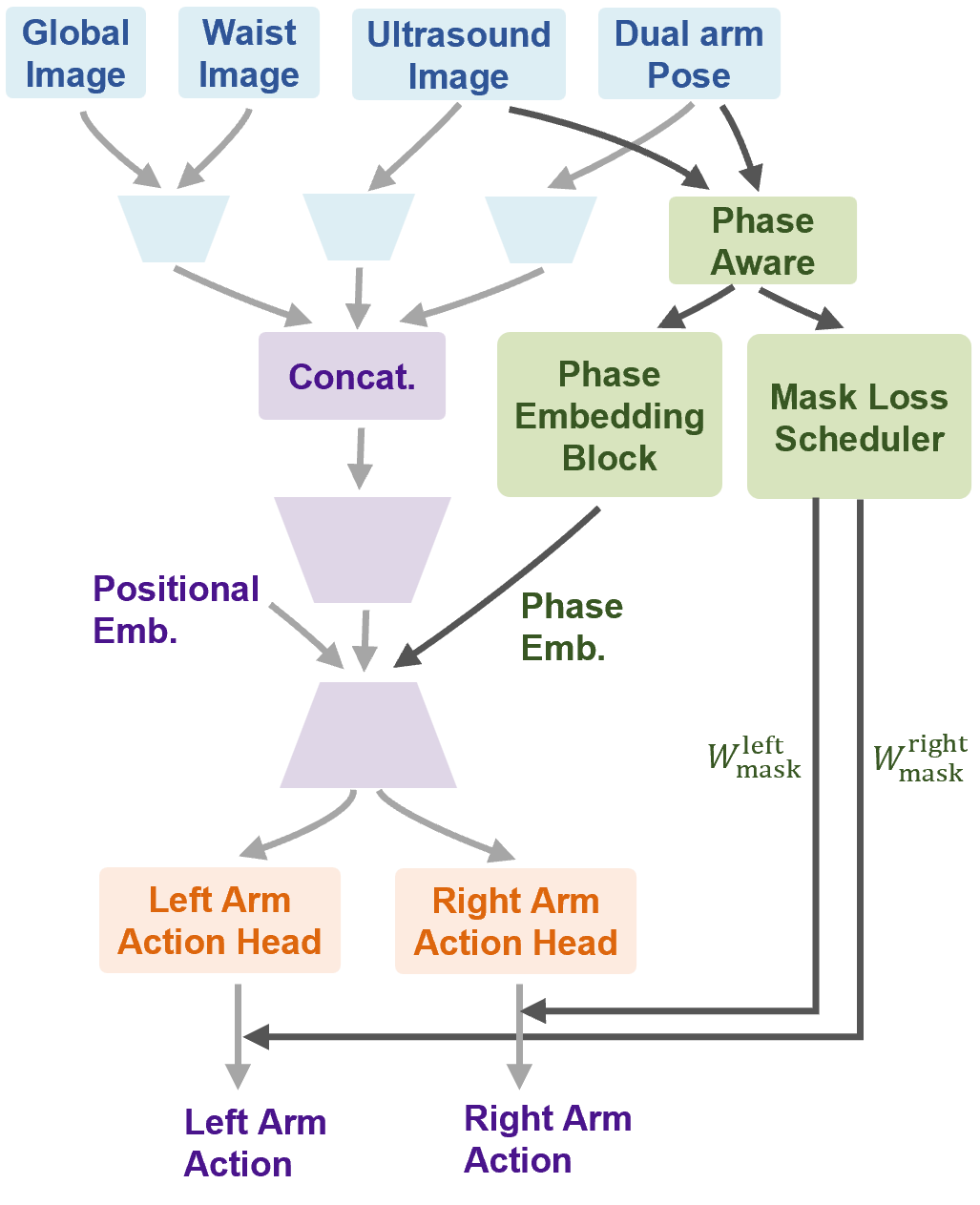}
\caption{\textbf{Phase-Aware Imitation Learning Module.} Network module color scheme: multimodal inputs (blue), transformer-based action chunking (purple), phase-aware module (green), and dual-arm outputs (orange) controlling the FR3 (left) and Panda (right) manipulators.}
\label{fig:network}
\end{figure}

The detailed architecture of our phase-aware imitation learning network is illustrated in Fig.~\ref{fig:network}. The network processes multimodal sensory data through three distinct pathways: (1) external visual feedback from wrist and overhead cameras; (2) internal anatomical feedback via a standalone ultrasound encoder; and (3) proprioceptive states of dual-arm poses. To navigate the high-variance clinical workflow, we introduce a phase-aware module that monitors ultrasound and proprioceptive streams. This module is triggered by specific state-based conditions to transition between operational phases, projecting this temporal logic into a phase embedding. Crucially, this embedding guides a mask loss scheduler, which superimposes pre-defined weights onto each specific phase. By applying these differentiated "influence weights," the scheduler dynamically modulates the learning objectives to prioritize phase-specific features. Finally, the network utilizes dual decoupled heads to output independent trajectories for the probe and needle, accommodating the task’s inherent asymmetry.

As illustrated in Fig.~\ref{fig:network}, the network processes multimodal data via three distinct encoders: a shared image encoder for external cameras, a dedicated ultrasound encoder, and a proprioceptive state encoder. To manage the clinical workflow's high temporal variance, a phase-aware module continuously monitors the ultrasound and proprioceptive streams, autonomously triggering phase transitions upon detecting specific multimodal conditions (e.g., target plane acquisition). This discrete phase state is projected into an embedding and injected into the Transformer decoder. During training, a mask loss scheduler superimposes pre-defined penalty weights onto each phase to dynamically modulate the optimization landscape. This explicitly forces the network to exhibit phase-appropriate behaviors, balancing smooth macro-transit with precise micro-manipulation. Finally, decoupled dual-action heads output independent trajectories for the asymmetric probe and needle control.

\section{EXPERIMENTS}

To comprehensively assess the proposed DAISS platform, our experimental validation is structured into two primary phases: a system-level feasibility analysis and an evaluation of the imitation learning policy. In the feasibility analysis, we rigorously validate the hardware and control framework by quantifying the operational safety of the bimanual workspace, the intrinsic precision of the optical tracking, and the end-to-end kinematic tracking accuracy during teleoperation.

\subsection{System Feasibility Evaluation}
Given the inherent complexity of the integrated system, a feasibility evaluation is essential. In particular, the overlapping workspaces of the dual-arm configuration necessitate stringent motion constraints to ensure seamless coordination. Consequently, we conducted a targeted assessment focusing on two critical performance metrics: operational safety and motion tracking accuracy

\subsubsection{Safety evaluation}

\begin{figure}
    \centering
    \includegraphics[width=0.4\textwidth]{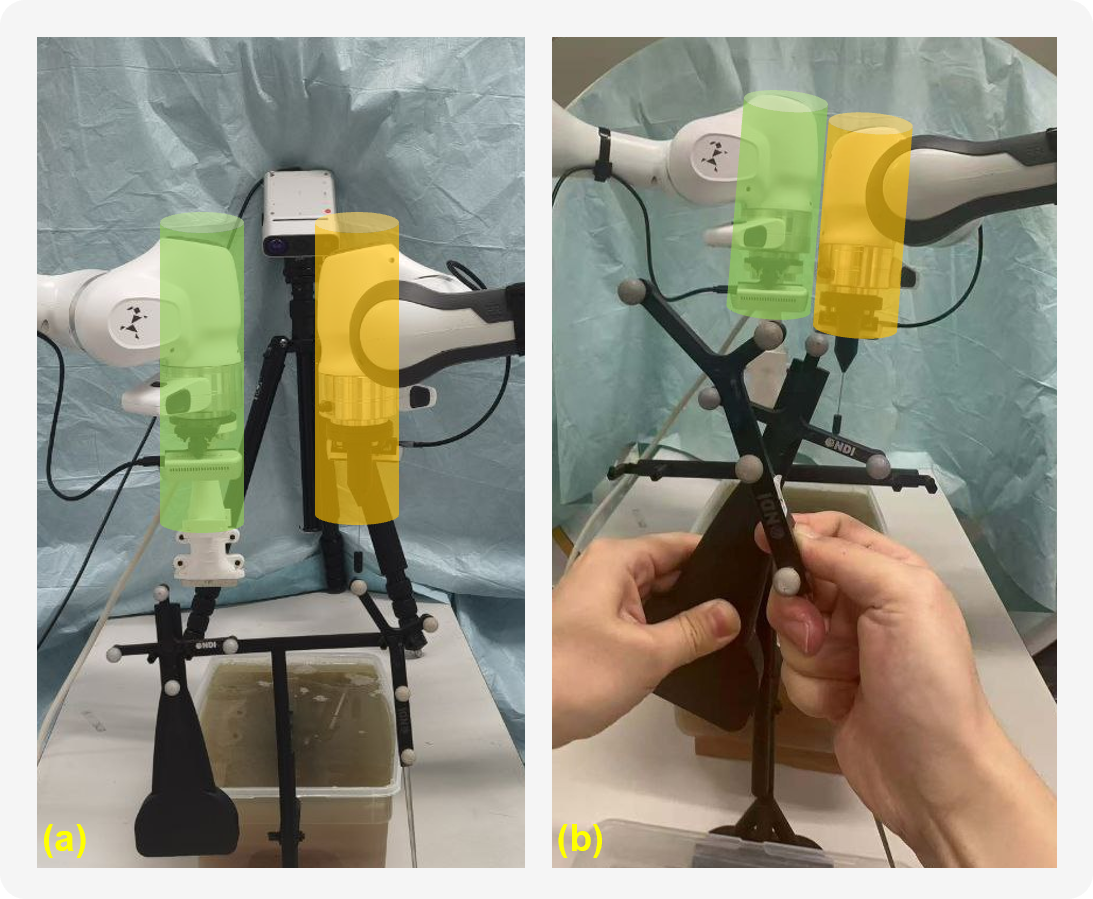}
    \caption{\textbf{Kinematic collision avoidance}. Green and yellow virtual cylinders represent the geometric envelopes of the robotic end-joints and tools. By acting as impenetrable spatial barriers, this setup inherently precludes mechanical interference between the dual manipulators during teleoperated maneuvers.}
    \label{fig:safety}
\end{figure}

Given the high dexterity and extensive range of motion of human hands, mapping their trajectories to robotic manipulators necessitates a robust consideration of workspace constraints. To account for the physical volume of the end effectors and their associated payloads, a virtual bounding cylinder is established at each TCP (Tool Center Point). This spatial envelope functions as a proactive collision-avoidance buffer, effectively mitigating inter-arm interference. The system’s operational reliability was evaluated across various motion patterns, as illustrated in Fig.~\ref{fig:safety}. Experimental results demonstrate that even during complex crossover maneuvers of the hand-held tools, the collision-avoidance mechanism reliably triggers a safety halt to prevent physical contact.

\subsubsection{Tracking evaluation}

\begin{figure}
    \centering
    \includegraphics[width=0.4\textwidth]{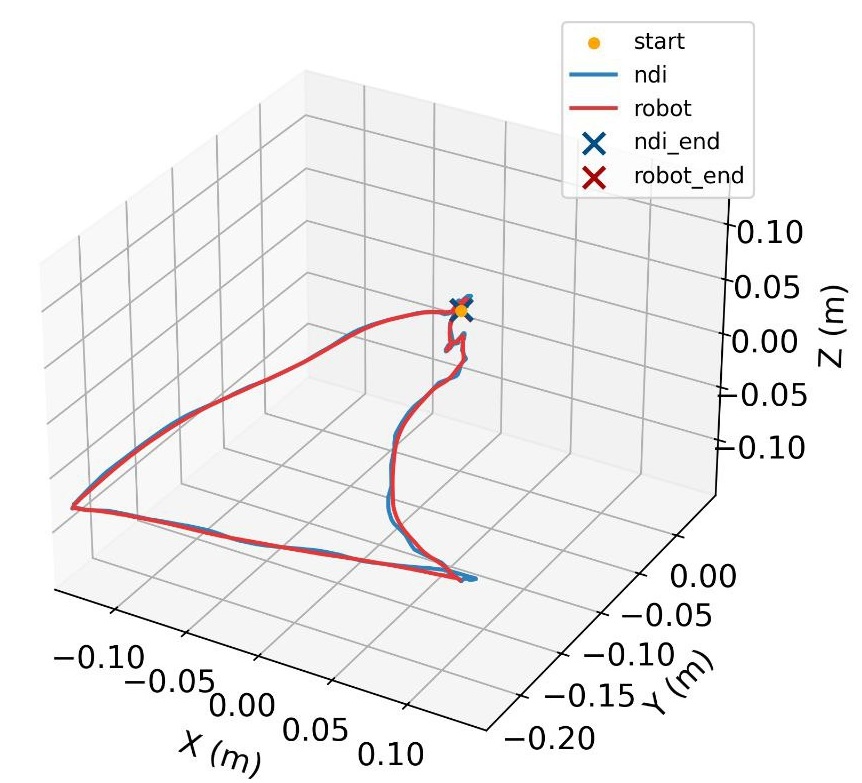}
    \caption{\textbf{Translational Tracking Performance}. }
    \label{fig:translation}
\end{figure}

\begin{figure}
    \centering
    \includegraphics[width=0.4\textwidth]{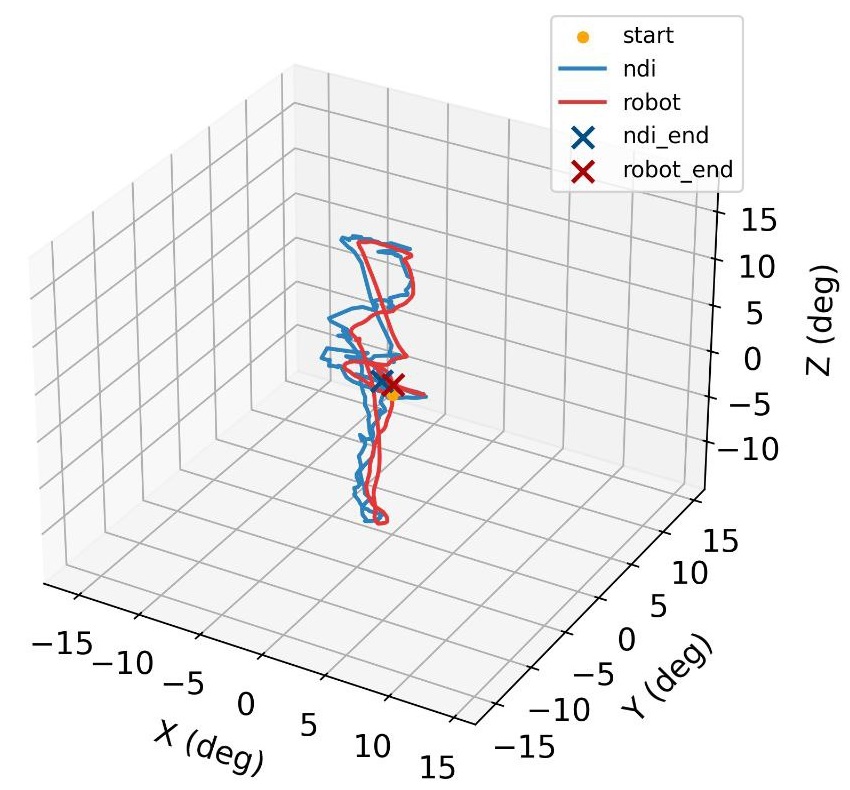}
    \caption{\textbf{Rotational Tracking Performance}. }
    \label{fig:rotation}
\end{figure}

To evaluate the fidelity of motion transfer from the NDI tracking system to the robotic manipulators, a comprehensive comparative analysis was conducted between the commanded trajectories derived from the tracking system and the actual measured states of the robot. To ensure rigorous temporal alignment, both data streams were synchronized via a high-frequency clock and sampled at a uniform rate, minimizing any phase shift during real-time data acquisition.
Each experimental trial followed a structured "move-and-hold" protocol: an operator manipulated the tracked tool from a predefined starting pose to a specific target pose, maintaining the final position for a two-second dwell time. This period allowed the robotic system to reach a steady-state condition and ensured that any transient oscillations had subsided before final measurements were taken.
Two primary metrics were employed to quantify system performance:
(1)Latency: Defined as the temporal delay between the moment the tracked tool arrived at its target pose and the moment the robotic manipulator reached its stabilized steady-state pose.
(2)Tracking Accuracy: Assessed by calculating the Euclidean distance for positional error and the geodesic distance for orientational error between the commanded target pose and the final measured robot state.
The quantitative results demonstrate the high precision of the system. The final steady-state positional error was measured at $0.622 \pm 0.478$ mm, while the orientational error remained as low as $0.222 \pm 0.095^\circ$. The mean system latency at the endpoints was recorded at $0.263 \pm 0.16$s. These values confirm that the system meets the stringent requirements for high-fidelity motion replication. The dynamic performance is further illustrated in Fig.~\ref{fig:translation} and Fig.~\ref{fig:rotation}, which display the tracking trajectories for translation and rotation, respectively. The data indicate that the robotic end-effectors follow translational commands with high fidelity. However, a slight discrepancy was observed in the rotational domain; specifically, high-frequency, low-amplitude rotations—occurring when the operator moved the tool "quickly and gently"—were occasionally filtered out. This phenomenon is likely attributed to the internal damping characteristics of the robot's motion controller and the low-pass filtering applied to the tracking signal to ensure smooth operation. Despite this minor filtering effect, the overall trajectory mapping remains highly consistent. The experimental evidence confirms that the system possesses significant robustness and feasibility for complex, coordinated dual-arm tasks, maintaining safety and precision even under dynamic manual guidance.

\subsection{Analysis of the Phase-Aware Imitation Learning}

To evaluate the efficacy of our phase-aware imitation learning policy, we decompose the complex ultrasound-guided interventional workflow into four distinct, sequentially dependent phases. This strategic decoupling is driven by multimodal state transitions, leveraging both robotic kinematics and real-time ultrasound image dynamics as natural phase triggers:

\textbf{Phase I—Probe Gross Positioning}. The FR3 manipulator navigates to the initial contact point on the phantom surface. This initialization is activated by a predefined system-start command.

\textbf{Phase II—Anatomical Scanning}. The FR3 performs an active sweep to locate the target acoustic window. This phase is autonomously triggered by the extraction of valid ultrasound image features, signifying established probe-tissue contact.

\textbf{Phase III—Needle Pre-alignment}. The Panda manipulator transits to the designated insertion site. Transition to this phase is strictly conditioned upon the stabilization of the target ultrasound plane achieved in Phase II.

\textbf{Phase IV—Fine-Grained Insertion}. The Panda executes the delicate needle insertion. This critical interaction is enabled by the global base-to-base calibration of the dual-arm system and is triggered by the spatial convergence of the ultrasound imaging plane and the needle trajectory.

To accommodate the divergent kinematic requirements of each stage, we introduce a dynamic mask loss that adaptively modulates the network's optimization landscape. During transit stages (Phases I and III), the objective is to generate smooth, stable macro-trajectories for efficient site reaching. Conversely, the interactive stages (Phases II and IV) demand intricate sensorimotor coordination for fine-tuning the ultrasound plane and needle angle. To prioritize these critical micro-manipulations, the dynamic mask assigns significantly higher penalty weights to Phases II and IV. This phase-conditioned weighting compels the imitation learning policy to allocate greater representational capacity to high-precision segments, ensuring responsiveness to subtle multimodal feedback.

To assess the proposed policy, we categorize the evaluation metrics into two dimensions: temporal efficiency and fineness of the action. For temporal efficiency, we focus specifically on Phase II (Anatomical Scanning), as it is most sensitive to ultrasound image feedback. We evaluated five distinct mask loss weights ($w \in \{0.1, 0.5, 1, 2, 10\}$). For each configuration, four trials were conducted, measuring the duration from initial ultrasound contact to target plane acquisition. The results are summarized in Table \ref{tab:temporal_efficiency}. 

\begin{table}[h]
\caption{Temporal efficiency with different weights for phase II}
\label{tab:temporal_efficiency}
\begin{center}
\begin{tabular}{cccc}
\toprule
\textit{Weight} & \textbf{0.1} & \textbf{0.5} & \textbf{1}  \\
\midrule
\textit{Time(s)} & $8.64 \pm 1.15$ & $12.44 \pm 0.61$ & $11.79 \pm 0.70$  \\
\toprule
\textit{Weight} & \textbf{2} & \textbf{10} \\
\midrule
\textit{Time(s)} & $15.18 \pm 3.30$ & $25.18 \pm 4.52$ \\
\bottomrule
\end{tabular}
\end{center}
\end{table}

\begin{figure}
    \centering
    \includegraphics[width=0.43\textwidth]{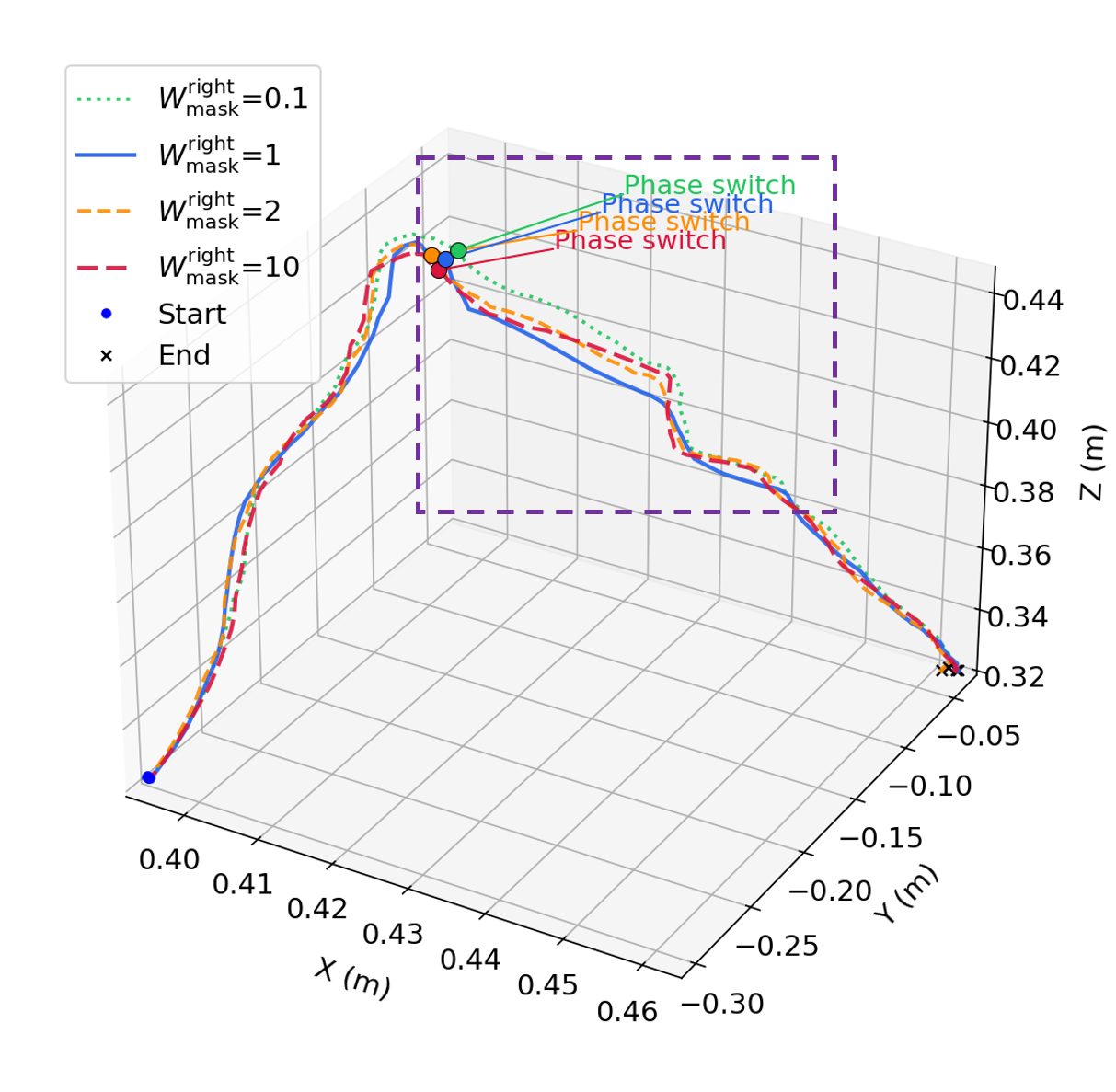}
    \caption{\textbf{Kinematic trajectories across varying weights in Phase IV.} The phase switch marker denotes the start of the final intervention stage.}
    \label{fig:ablation_trajectory}
\end{figure}

Although configuring the phase weight to 0.1 yields the shortest execution time, it consistently suffers from severe trajectory overshooting. To systematically investigate this inherent speed-accuracy trade-off, we further evaluate the kinematic precision of the robotic actions. To quantify the spatial fidelity of the micro-manipulations, Fig.~\ref{fig:ablation_trajectory} illustrates the impact of varying the dynamic mask weights during Phase IV. As highlighted in the purple dashed box (corresponding to the fine-grained insertion stage), the trajectory profile becomes progressively more deliberate and defined as the weight increases from 0.1 to 10. This steepening effect physically demonstrates that higher penalty weights successfully force the policy to capture the subtle, high-frequency spatial nuances of the expert demonstrations, effectively preventing the network from generating overly smoothed, low-precision paths.

\begin{figure}
    \centering
    \includegraphics[width=0.45\textwidth]{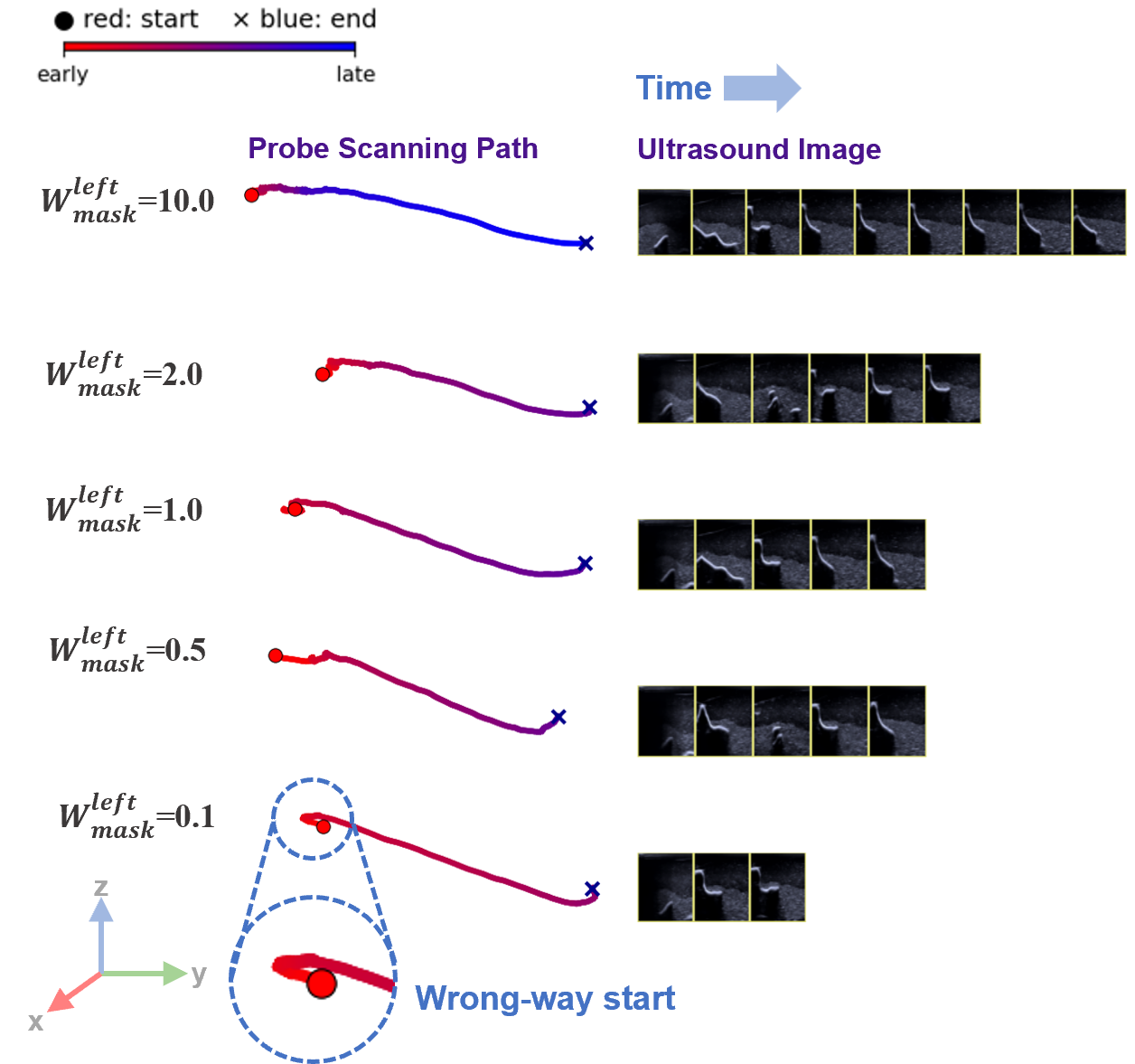}
    \caption{\textbf{Phase II trajectory ablation}. Kinematic execution of the anatomical scanning phase across varying weight parameters.}
    \label{fig:ablation}
\end{figure}

As illustrated in Fig.~\ref{fig:ablation}, we conducted an ablation study to evaluate the impact of varying the dynamic mask weights during Phase II (Anatomical Scanning). As the penalty weight increases from 0.1 to 10.0, the robotic scanning speed decreases progressively (from red to blue), allowing the policy to allocate greater attention to the fine-grained anatomical details within the ultrasound stream. However, this introduces a critical trade-off. Assigning an insufficiently low weight leads to overly aggressive sweeping motions, rendering the system prone to erroneous state transitions and missed target planes. Conversely, an excessively high weight induces overly conservative kinematics, causing the scanning motion to stall and ultimately fail to reach the desired clinical endpoint.

\section{DISCUSSION}
Compared to standard robotic manipulation, dual-arm clinical interventions introduce significantly more stringent constraints. The system must coordinate asymmetric payloads, such as a bulky ultrasound probe and a delicate needle, while executing high-accuracy trajectories within a highly confined, shared workspace. To overcome these challenges, the proposed DAISS platform seamlessly fuses real-time ultrasound streams, external camera vision, and robotic kinematics into a comprehensive multimodal proprioceptive framework. By leveraging this rich sensory representation, our phase-aware imitation learning policy successfully captures and distills expert surgical intuition in a data-efficient manner, demonstrating strong potential not only for autonomous intervention but also for intelligent novice training.

A core conceptual advantage of our approach is the explicit decoupling of the operation workflow. Rather than forcing a single network to compromise between speed and precision, the dynamic mask loss adaptively shifts the optimization landscape for each distinct phase. It prioritizes smooth, rapid macro-level transit during the approach stages, while forcing the policy to slow down and heavily weight ultrasound image features during critical micro-manipulation stages. As a result, the pervasive issues of kinematic overshoot and high-frequency jitter were effectively resolved, perfectly balancing the inherent speed-accuracy trade-off without significantly sacrificing overall execution time.

Furthermore, safety remains the paramount priority in such setups. Our current collision avoidance mechanism—geometrically abstracting the end-effectors as virtual spatial cylinders—provides a robust, computationally lightweight kinematic safeguard. Moving forward, this geometric approach can be further augmented by embedding spatial awareness directly into the neural policy, enabling proactive, learning-based obstacle avoidance. Finally, while our phantom-based experiments rigorously validate the multimodal fusion and kinematic accuracy of the DAISS platform, clinical realities present exciting avenues for future refinement. In vivo environments naturally introduce dynamic soft-tissue deformations, such as respiratory motion, which will require advanced temporal tracking in future iterations. Additionally, while our vision-proprioception fusion ensures spatial precision, integrating force-torque sensors for active impedance control would add a vital tactile dimension, ensuring optimal and safe tissue-contact dynamics during continuous scanning.

\section{CONCLUSIONS}
In this paper, we introduced DAISS, an advanced dual-arm teleoperation platform designed to capture high-fidelity, multimodal expert demonstrations for ultrasound-guided interventions. Building upon this hardware framework, we proposed a novel phase-aware imitation learning policy, optimized via a dynamic mask loss, to accurately distill intricate expert surgical intuition. Experimental evaluations demonstrate that our approach successfully decouples the complex clinical workflow, effectively balancing temporal efficiency with fine-grained kinematic precision. By adaptively modulating optimization weights during critical micro-manipulation stages, the proposed policy eliminates the severe trajectory overshooting commonly observed in conventional methods. Ultimately, this work establishes a robust algorithmic and hardware foundation for achieving safe, autonomous bimanual interventions in dynamic and asymmetric clinical environments.

\bibliographystyle{IEEEtran}
\bibliography{bibliography}

@article{lorentzen2015efsumb,
  title={EFSUMB guidelines on interventional ultrasound (INVUS), part I--general aspects (long version)},
  author={Lorentzen, Torben and Nols{\o}e, CP and Ewertsen, C and Nielsen, Michael B and Leen, Edward and Havre, Roland F and Gritzmann, N and Brkljacic, B and N{\"u}rnberg, D and Kabaalioglu, Adnan and others},
  journal={Ultraschall in der Medizin-European Journal of Ultrasound},
  volume={36},
  number={05},
  pages={E1--E14},
  year={2015},
  publisher={{\copyright} Georg Thieme Verlag KG}
}

@article{raitor2023design,
  title={Design and Evaluation of Haptic Guidance in Ultrasound-Based Needle-Insertion Procedures},
  author={Raitor, Michael and Nunez, Cara M and Stolka, Philipp J and Okamura, Allison M and Culbertson, Heather},
  journal={IEEE Transactions on Biomedical Engineering},
  volume={71},
  number={1},
  pages={26--35},
  year={2023},
  publisher={IEEE}
}

@article{li2025robotic,
  title={Robotic CBCT meets robotic ultrasound},
  author={Li, Feng and Bi, Yuan and Huang, Dianye and Jiang, Zhongliang and Navab, Nassir},
  journal={International Journal of Computer Assisted Radiology and Surgery},
  pages={1--9},
  year={2025},
  publisher={Springer}
}

@inproceedings{jiang2021motion,
  title={Motion-aware robotic 3D ultrasound},
  author={Jiang, Zhongliang and Wang, Hanyu and Li, Zhenyu and Grimm, Matthias and Zhou, Mingchuan and Eck, Ulrich and Brecht, Sandra V and Lueth, Tim C and Wendler, Thomas and Navab, Nassir},
  booktitle={2021 IEEE International Conference on Robotics and Automation (ICRA)},
  pages={12494--12500},
  year={2021},
  organization={IEEE}
}

@article{li2021overview,
  title={An overview of systems and techniques for autonomous robotic ultrasound acquisitions},
  author={Li, Keyu and Xu, Yangxin and Meng, Max Q-H},
  journal={IEEE Transactions on Medical Robotics and Bionics},
  volume={3},
  number={2},
  pages={510--524},
  year={2021},
  publisher={IEEE}
}

@inproceedings{song2025intelligent,
  title={Intelligent Virtual Sonographer (IVS): Enhancing Physician-Robot-Patient Communication},
  author={Song, Tianyu and Li, Feng and Bi, Yuan and Karlas, Angelos and Yousefi, Amir and Branzan, Daniela and Jiang, Zhongliang and Eck, Ulrich and Navab, Nassir},
  booktitle={International Conference on Medical Image Computing and Computer-Assisted Intervention},
  pages={287--297},
  year={2025},
  organization={Springer}
}

@article{jiang2022towards,
  title={Towards autonomous atlas-based ultrasound acquisitions in presence of articulated motion},
  author={Jiang, Zhongliang and Gao, Yuan and Xie, Le and Navab, Nassir},
  journal={IEEE Robotics and Automation Letters},
  volume={7},
  number={3},
  pages={7423--7430},
  year={2022},
  publisher={IEEE}
}

@article{kojcev2016dual,
  title={Dual-robot ultrasound-guided needle placement: closing the planning-imaging-action loop},
  author={Kojcev, Risto and Fuerst, Bernhard and Zettinig, Oliver and Fotouhi, Javad and Lee, Sing Chun and Frisch, Benjamin and Taylor, Russell and Sinibaldi, Edoardo and Navab, Nassir},
  journal={International journal of computer assisted radiology and surgery},
  volume={11},
  number={6},
  pages={1173--1181},
  year={2016},
  publisher={Springer}
}

@inproceedings{koskinopoulou2023dual,
  title={Dual robot collaborative system for autonomous venous access based on ultrasound and bioimpedance sensing technology},
  author={Koskinopoulou, Maria and Acemoglu, Alperen and Penza, Veronica and Mattos, Leonardo S},
  booktitle={2023 IEEE International Conference on Robotics and Automation (ICRA)},
  pages={4648--4653},
  year={2023},
  organization={IEEE}
}

@article{li2025ultrasound,
  title={An ultrasound visual servoing dual-arm robotics system for needle placement in brachytherapy treatment},
  author={Li, Yanlei and Lu, Zhenyu and Tzemanaki, Antonia and Bahl, Amit and Persad, Raj and Melhuish, Chris and Yang, Chenguang},
  journal={Frontiers in Robotics and AI},
  volume={12},
  pages={1558182},
  year={2025},
  publisher={Frontiers Media SA}
}

@article{zhao2023learning,
  title={Learning fine-grained bimanual manipulation with low-cost hardware},
  author={Zhao, Tony Z and Kumar, Vikash and Levine, Sergey and Finn, Chelsea},
  journal={arXiv preprint arXiv:2304.13705},
  year={2023}
}

@article{zhao2024aloha,
  title={Aloha unleashed: A simple recipe for robot dexterity},
  author={Zhao, Tony Z and Tompson, Jonathan and Driess, Danny and Florence, Pete and Ghasemipour, Kamyar and Finn, Chelsea and Wahid, Ayzaan},
  journal={arXiv preprint arXiv:2410.13126},
  year={2024}
}

@article{kim2024surgical,
  title={Surgical robot transformer (srt): Imitation learning for surgical tasks},
  author={Kim, Ji Woong and Zhao, Tony Z and Schmidgall, Samuel and Deguet, Anton and Kobilarov, Marin and Finn, Chelsea and Krieger, Axel},
  journal={arXiv preprint arXiv:2407.12998},
  year={2024}
}

@article{kim2025srt,
  title={SRT-H: A hierarchical framework for autonomous surgery via language-conditioned imitation learning},
  author={Kim, Ji Woong and Chen, Juo-Tung and Hansen, Pascal and Shi, Lucy Xiaoyang and Goldenberg, Antony and Schmidgall, Samuel and Scheikl, Paul Maria and Deguet, Anton and White, Brandon M and Tsai, De Ru and others},
  journal={Science robotics},
  volume={10},
  number={104},
  pages={eadt5254},
  year={2025},
  publisher={American Association for the Advancement of Science}
}

@inproceedings{chen2025ultradp,
  title={UltraDP: Generalizable Carotid Ultrasound Scanning with Force-Aware Diffusion Policy},
  author={Chen, Ruoqu and Yan, Xiangjie and Lv, Kangchen and Huang, Gao and Li, Zheng and Li, Xiang},
  booktitle={2025 IEEE/RSJ International Conference on Intelligent Robots and Systems (IROS)},
  pages={20074--20080},
  year={2025},
  organization={IEEE}
}

@article{jiang2024intelligent,
  title={Intelligent robotic sonographer: Mutual information-based disentangled reward learning from few demonstrations},
  author={Jiang, Zhongliang and Bi, Yuan and Zhou, Mingchuan and Hu, Ying and Burke, Michael and Navab, Nassir},
  journal={The International Journal of Robotics Research},
  volume={43},
  number={7},
  pages={981--1002},
  year={2024},
  publisher={SAGE Publications Sage UK: London, England}
}

@article{horaud1995hand,
  title={Hand-eye calibration},
  author={Horaud, Radu and Dornaika, Fadi},
  journal={The international journal of robotics research},
  volume={14},
  number={3},
  pages={195--210},
  year={1995},
  publisher={Sage Publications Sage CA: Thousand Oaks, CA}
}

\end{document}